\documentclass[superscriptaddress,amsmath,twocolumn,amssymb,preprint,10pt]{revtex4-2}

\usepackage{graphicx} % Required for the inclusion of images
\usepackage{amsmath} % Required for some math elements
\usepackage{xifthen}
\usepackage{physics}
\usepackage{hyperref}
\usepackage{etoolbox}
\usepackage{tabularx}
\usepackage{soul, todonotes}
\soulregister\cite7
\soulregister\ref7

\patchcmd{\section}
  {\centering}
  {\raggedright}
  {}
  {}
\patchcmd{\subsection}
  {\centering}
  {\raggedright}
  {}
  {}

\newcommand{\nn}[2]{^{(#1)}_{#2}}
\newcommand{\Loss}{\mathcal{L}}

\begin{document}

\title{Hybrid training of optical neural networks}

\author{James Spall}
\thanks{These authors contributed equally to this work.}
\affiliation{Clarendon Laboratory, University of Oxford, Parks Road, Oxford,  OX1 3PU, UK}

\author{Xianxin Guo}
\thanks{These authors contributed equally to this work.}
\email{xianxin.guo@physics.ox.ac.uk}
\affiliation{Clarendon Laboratory, University of Oxford, Parks Road, Oxford,  OX1 3PU, UK}
\affiliation{OxONN, Wood Centre for Innovation, Quarry Road, Headington, Oxford, OX3 8SB, UK}

\author{A. I. Lvovsky}
\email{alex.lvovsky@physics.ox.ac.uk}
\affiliation{Clarendon Laboratory, University of Oxford, Parks Road, Oxford,  OX1 3PU, UK}
\affiliation{OxONN, Wood Centre for Innovation, Quarry Road, Headington, Oxford, OX3 8SB, UK}
%\affiliation{\hl{Russian Quantum Center, Skolkovo, 143025, Moscow, Russia}}

\date{\today}

\begin{abstract}
Optical neural networks are emerging as a promising type of machine learning hardware capable of energy-efficient, parallel computation. Today's optical neural networks are mainly developed to perform optical inference after \emph{in silico} training on digital simulators. However, various physical imperfections that cannot be accurately modelled may lead to the notorious ``reality gap'' between the digital simulator and the physical system. To address this challenge, we demonstrate hybrid training of optical neural networks where the weight matrix is trained with neuron activation functions computed optically via forward propagation through the network.
We examine the efficacy of hybrid training with three different networks: an optical linear classifier, a hybrid opto-electronic network, and a complex-valued optical network. We perform a comparative study to \emph{in silico} training, and our results show that hybrid training is robust against different kinds of static noise. Our platform-agnostic hybrid training scheme can be applied to a wide variety of optical neural networks, and this work paves the way towards advanced all-optical training in machine intelligence.
\end{abstract}

\maketitle

\section{Introduction}

Machine learning powered by artificial neural networks has reshaped the landscape in many different areas over the last decade. This machine learning revolution is fuelled by the immense parallel computing power of electronic hardware such as graphics- and tensor- processing units. However, the rapid growth of computational demand in this field has outpaced Moore's law, and today's machine learning applications are associated with high energy cost and terrible carbon footprint ~\cite{patterson2021carbon}. We are in dire need of novel computing systems capable of fast and energy-efficient computation to drive the future development of machine learning.

Optics provides a promising analog computing platform, and optical neural networks (ONNs) have recently been the focus of intense research and commercial interest. Thanks to the superposition and coherence properties of light, neurons in ONNs can be naturally connected via interference~\cite{shen2017deep, abu1987optical, farhat1985optical} or diffraction~\cite{lin2018all, zhou2021large, li2021spectrally} in different settings, whilst the neuron activation function can be physically implemented with a large variety of nonlinear optical effects~\cite{abu1987optical, cruz2000reinforcement, zuo2019all}. Together these resources have enabled the optical realization of various neural network architectures, including fully connected~\cite{shen2017deep, lin2018all, zhou2021large, li2021spectrally}, convolutional~\cite{xu202111, feldmann2021parallel, miscuglio2020massively} and recurrent~\cite{abu1987optical, farhat1985optical, bueno2018reinforcement, larger2012photonic}. Today's advanced optical technologies have already allowed ONNs to reach a computational speed of ten trillion operations per second~\cite{xu202111}, comparable to that of their electronic counterparts; and the energy consumption can be on a scale of, or even less than, one photon per operation~\cite{wang2021optical}, orders of magnitude lower than that of digital computation.

Current ONNs are primarily developed to perform inference tasks in machine learning~\cite{wetzstein2020inference, shastri2021photonics}, and they are usually trained on a digital computer. During this \emph{in silico} training, one has to simulate the physical system digitally, then apply the standard ``backpropagation'' algorithm~\cite{lecun2015deep}, which involves repeated forward- and backward- propagation of information inside the network. The update of the weight matrices is computed from the combined data obtained in these two processes. Because any physical system exhibits certain experimental imperfections that are hard to accurately model, ONNs trained in this way usually perform worse than expected~\cite{zhou2021large, zuo2019all, wright2022deep}. To narrow this reality gap, one can incorporate simulated noise into the \emph{in silico} training~\cite{li2021spectrally}. However this approach is suboptimal because it does not incorporate the specific pattern of imperfections that is present in a given ONN. \textcolor{black}{Another approach is to apply platform-specific error correction algorithms~\cite{miller2015perfect, bandyopadhyay2021hardware, miller2017setting, pai2020parallel, cramer2022surrogate} or iterative optimisation algorithms~\cite{gerchberg1972practical} to reduce experimental imperfections, but running these algorithms can be time and resource demanding.}

Our group has recently proposed a method for obtaining the training signal directly from the optical fields propagating through the network in both directions~\cite{guo2021backpropagation}. %This method allows one to not only train faster, but also build the actual physics of the system, including its imperfections, into the training, thereby largely closing the reality gap. 
This method not only allows faster training, but also helps close the reality gap, as the actual physics of the system, including its imperfections, is built into the training. 

\begin{figure*}[t]
		\centering
		\includegraphics[width=\textwidth]{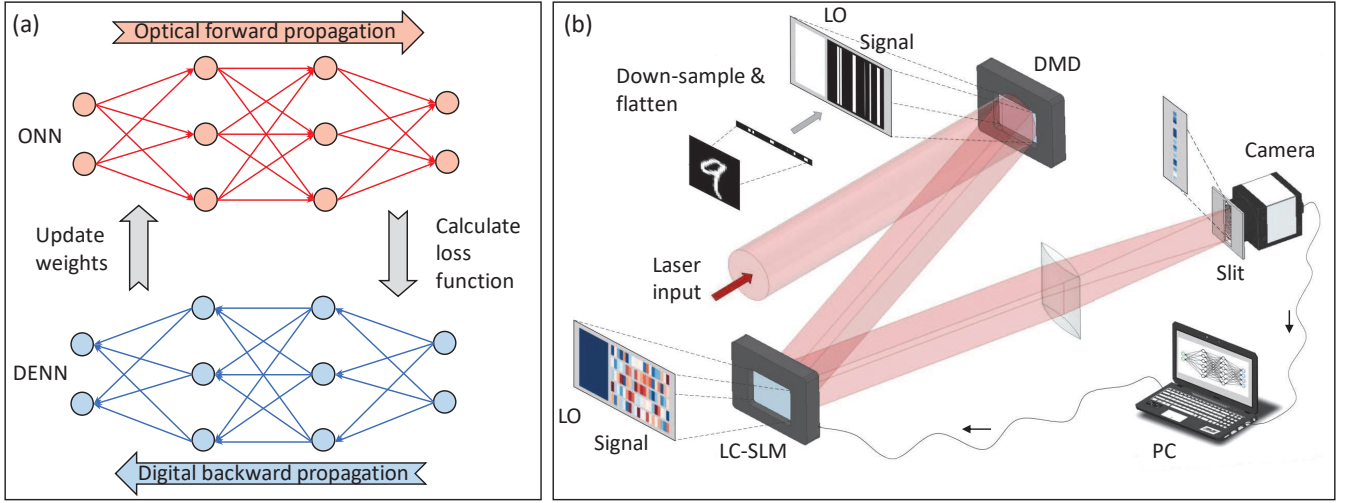}
		\caption{\textbf{Hybrid training of ONNs.} (a) Conceptual illustration of the scheme, where forward propagation is implemented in the optical system and error backpropagation is implemented digitally. (b) Schematic of our ONNs built upon a single-layer optical multiplier. }
		\label{fig: schematic}
	\end{figure*}
	
In this work, we demonstrate the first step towards this vision. In our training scheme, the forward \textcolor{black}{linear computation} and backward propagation of information is implemented with optics and electronics, respectively. The portion of the training signal that is acquired from forward propagation is then obtained through direct measurement of the optical field \textcolor{black}{followed by digital calculation of the nonlinear activation function}. 

% Although this hybrid training method has been demonstrated in analog neural networks or neuromorphic computing platforms based on memresistors~\cite{mennel2020ultrafast, li2018efficient, wang2019situ}, it hasn't been fully implemented or systematically studied in the optical domain. Recently Wright {\it et al.}~\cite{wright2022deep} demonstrated ``Physics-Aware Training'' of various physical networks, of which one was opto-electronic.  %network with electronic linear layers and optical nonlinearity. 

\textcolor{black}{This hybrid training method, also known as ``hardware-in-the-loop training", is a well established technique that incorporates the actual hardware response in each update loop, to mitigate the impact of hardware imperfections, and has been demonstrated in various electronic analog neural networks as well as neuromorphic computing platforms based on memristors~\cite{tam1990learning, mennel2020ultrafast, li2018efficient, wang2019situ}. Recently Wright \textit{et al.}~\cite{wright2022deep} demonstrated ``physics-aware training'' of an opto-electronic network. However, in this case the optical system did not contain a fully tunable weight matrix, so an additional fully connected digital linear layer had to be included.}
%only the activation of the neurons in the forward propagation was implemented by means of optical nonlinearity, whereas both the linear portion of the forward propagation, as well as the entire backward propagation, was performed electronically. 
Since the computational advantages of ONNs reside in optical linear layers \textcolor{black}{with controllable weight matrices}, and experimental imperfections often originate from the optical linear connection, optical propagation through the entire network is required for proper hybrid training.

Another related work is that of Zhou {\it et al.}~\cite{zhou2021large}. After training a 3-layer ONN \emph{in silico}, the authors tested their network optically, made corrections to the digital models of the second and third layers to account for the measured performance of the first layer, and digitally re-trained these layers. The optical testing and re-training was then repeated just for the third layer. \textcolor{black}{While this is a step towards the hybrid training scheme, a major shortcoming of this approach is the requirement of initial \emph{in silico} training. Another shortcoming is that the behavior of later layers must be well characterized and modelled to compensate for the errors caused by previous layers.} Furthermore, the training must be repeated multiple times.
 
In this work we demonstrate hybrid training on an ONN platform with a fully-connected layer. Full connectivity provides compatibility with digital neural network architectures and hence a degree of universality that is not offered by other architectures \textcolor{black}{such as reservoir computing, kernel machines and spiking neural networks.} We construct three different ONNs: an optical linear classifier, a hybrid neural network with optical and electronic layers, and a complex-valued ONN. Our work is \textcolor{black}{different from} that of Wright {\it et al.}, because all these networks contain optical linear layers. And unlike Zhou {\it et al.}, initial \emph{in silico} training is not required, and the optical signal is used to compute every single update of the physical weight matrices. We further analyze the performance of our networks under the influence of various types of noise and find significant improvement in comparison to \emph{in silico} training provided that the noise is static, i.e.~does not change in time.

\section{Results}
\subsection{Network training}
In supervised learning, the weight matrices of a neural network are iteratively updated (trained) via the ``backpropagation'' algorithm~\cite{lecun2015deep}, eventually enabling the network to replicate the mapping between the network input and the \textcolor{black}{ground truth}. The training is implemented using a labelled dataset $(\textbf{x}, \textbf{t})$, where $\textbf{x}$ is sent to the network input ($a^{(0)}_{i}=x_{i}$), and $\textbf{t}$ is the label to be compared with the network output. The neurons in subsequent layers are interconnected as
\begin{equation}\label{eq:forward}
z\nn{l}{j} = \sum_{i} w\nn{l}{ji} a\nn{l-1}{i},
\end{equation}
where $a\nn{l}{j} = g(z\nn{l}{j})$ is the nonlinear activation of each neuron. A loss function, $\Loss$, is defined in order to quantify the divergence between the network output and the correct label. Its  gradient with respect to the weights is 
\begin{equation}
\frac{\partial \Loss}{\partial w\nn{l}{ji}}
= \frac{\partial \Loss}{\partial z\nn{l}{j}}\frac{\partial z\nn{l}{j}}{\partial w\nn{l}{ji}}
= \delta\nn{l}{j}a\nn{l-1}{i},
\label{eq:weight_grad}
\end{equation}
where $\delta\nn{l}{j} \equiv \partial \Loss/\partial z\nn{l}{j}$ is referred to as the ``error'' at the $j$-th neuron in the $l$-th layer. By applying the chain rule of calculus, we have
\begin{equation}
\delta\nn{l}{j}
= \sum_{k}\frac{\partial \Loss}{\partial z\nn{l+1}{k}}\frac{\partial z\nn{l+1}{k}}{\partial z\nn{l}{j}}
= g'(z\nn{l}{j})\rho_j^{(l+1)},
\label{eq:neuron_grad}
\end{equation}
where $\rho_j^{(l+1)}{=}\sum_{k}w\nn{l+1}{kj}\delta\nn{l+1}{k}$. From \eqref{eq:neuron_grad} we see that the error vector inside the network can be calculated from the error vector at the subsequent layer, and the error vector at the output layer is directly calculated from the loss function. Once these error vectors, as well as the activations $a\nn{l-1}{}$ of all neurons are known, the gradients (\ref{eq:weight_grad}) of the loss function with respect to all the weights can be calculated, and hence the weights can be iteratively updated via gradient descent until convergence. This standard procedure is efficient in training digital electronic neural networks (DENNs). To train an ONN, one can model the network architecture on the computer, and implement the backpropagation algorithm digitally. The final weights after the training are then transferred to the ONN to perform inference tasks. This is called the \emph{in silico} training method.

As we see from Eq.~\eqref{eq:weight_grad}, the gradient matrix in each layer is the outer product of the corresponding activation and error vectors. In our hybrid training scheme, the activation vectors are obtained through optical forward propagation of neuron values [Fig.~\ref{fig: schematic}(a, top panel)]. These values are measured by photodetectors and recorded digitally. The error vectors, on the other hand, are obtained through digital backpropagation [Fig.~\ref{fig: schematic}(a, bottom panel)]. 
 Once the weight gradients (\ref{eq:weight_grad}) are calculated, the physical weights of the ONNs are updated. This hybrid training process is repeated until convergence.

\begin{figure*}[t]
		\centering
		\includegraphics[width=\textwidth]{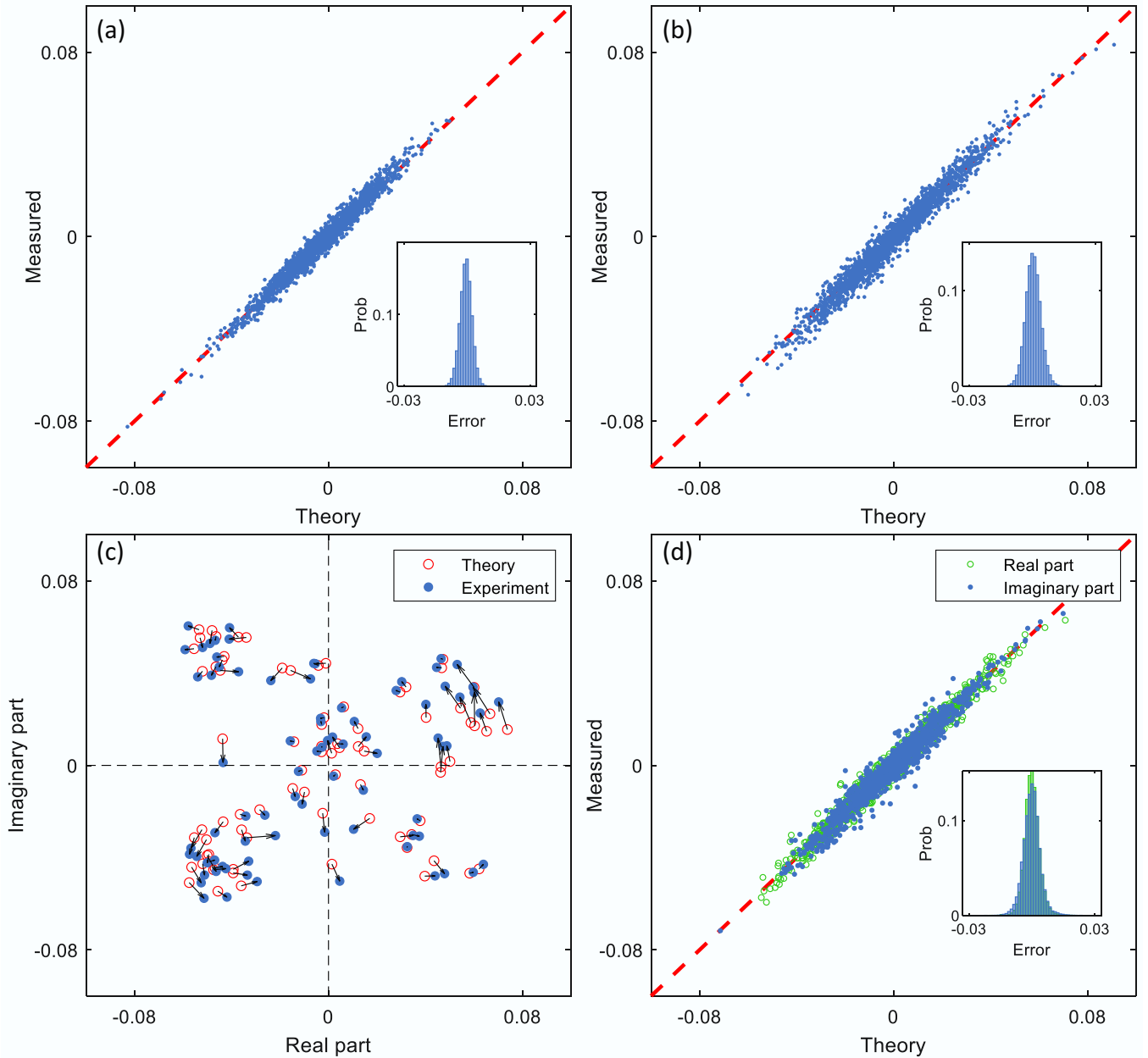}
		\caption{\textbf{Characterization of optical matrix-vector multiplication.} Performance of real-valued optical MVM with matrix size of (a) $100\times10$ and (b) $100\times25$. Ideal results would fall along the red dashed diagonal line. Distributions of the multiplication error are shown in insets. All the values are normalized by the maximum possible output. Because the weight matrix elements are normally distributed around zero, the vast majority of neuron values fall within a narrow range, also centered around zero. (c) Examples of the complex-valued optical MVM output (blue dots) compared with theory targets (red circles) on the complex plane. The differences are indicated by black arrows. (d) Performance of complex-valued optical MVM with matrix size of $100\times10$. Real and imaginary parts of the experimental result are shown in green circles and blue dots.}
		\label{fig: multiplier}
	\end{figure*}

\subsection{Experimental system}

In a neural network, the interconnection of neurons (\ref{eq:forward}) is achieved by matrix-vector multiplication (MVM), and this basic operation constitutes the major computational workload in machine learning. Here we build a reconfigurable, coherent optical multiplier with spatial light modulators and lenses, as shown in Fig.~\ref{fig: schematic}(b).

We encode the neuron values in the electric field amplitude of the light propagating through the network. The positive-valued input vector \textcolor{black}{is converted to a 4-bit integer and then} encoded in the spatial field distribution of the light by a digital micromirror device (DMD, Texas Instruments DLP 6500), which is a binary amplitude-only modulator. \textcolor{black}{Each input vector element is encoded in a rectangular pixel block, whose width is 15 pixels, and height is the entire vertical dimension of the DMD. The 4-bit vector element value is represented by the fraction of pixels that are switched ‘on’ in each block.} A liquid-crystal phase-only spatial light modulator (LC-SLM, Santec SLM 100, \textcolor{black}{10-bit}) is used to further encode an arbitrary real-valued or complex-valued weight matrix. To this end, we apply a phase grating pattern on the LC-SLM and adjust the local offset and height of the grating so as to obtain the required phase and amplitude pattern in the first diffraction order~\cite{spall2020fully, arrizon2007pixelated}. \textcolor{black}{With this encoding method we are also able to correct for various optical aberrations and the severe wavefront distortion from both SLMs.} As the light passes through the DMD and LC-SLM planes, each row of the weight matrix is multiplied in an element-wise fashion by the input vector, and the results are then summed by means of a cylindrical lens. The output is measured with a fast CMOS camera (Basler ace acA640-750um, \textcolor{black}{8-bit}). The actual setup includes several additional lenses (not shown in the simplified schematic of Fig.~\ref{fig: schematic}(b)) for the elimination of unwanted diffraction. Details of our encoding method and actual setup have been illustrated in our previous work~\cite{spall2020fully}.

To read out the neuron values encoded in the optical field amplitudes, we perform homodyne detection. This detection requires a local oscillator (LO), which must be phase-stable with respect to the signal. We address this requirement by  allocating part of the DMD and LC-SLM active area as the reference region, and keeping these pixels fully reflective. The portion of the beam that reflects from the reference region serves as the LO. Both the signal and LO fields propagate through the entire system, and so the cylindrical lens not only completes the MVM, but also mixes the LO field with the MVM result at the output plane. \textcolor{black}{For real-valued MVM, we set the LO phase equal to that of the signal. The intensity measured by the camera then equals $I=(E_{\rm LO}+E_s)^2$, where $E_{\rm LO}$ and $E_s$ are the amplitudes of the two fields, and we ensure that $E_{\rm LO}>E_s$. The signal field amplitude is then found according to $E_s=\sqrt I-E_{\rm LO}$.}  Since both the LO and signal share the same optical path together with all the optical elements, their relative phase barely fluctuates, and the phase offset can be conveniently set by the LC-SLM. Our homodyne detection scheme avoids the extra experimental complexity of introducing an external LO beam and actively stabilizing the relative phase.

% To readout the neuron values, we perform interferometric detection to measure the electric field. Instead of introducing an additional reference beam, we allocate part of the DMD and LC-SLM active area as the reference region, and keep these pixels fully reflective. Because both the signal and reference pass through the same cylindrical lens, they interfere at the output plane, and intensity readout directly reveals the electric field of the MVM signal. The reference and signal share the same optical path together with all the optical elements, so their relative phase shows only minor fluctuations, and the relative phase can be conveniently set by the LC-SLM.

\subsection{Real-valued ONNs}
\begin{figure*}[t]
		\centering
		\includegraphics[width=\textwidth]{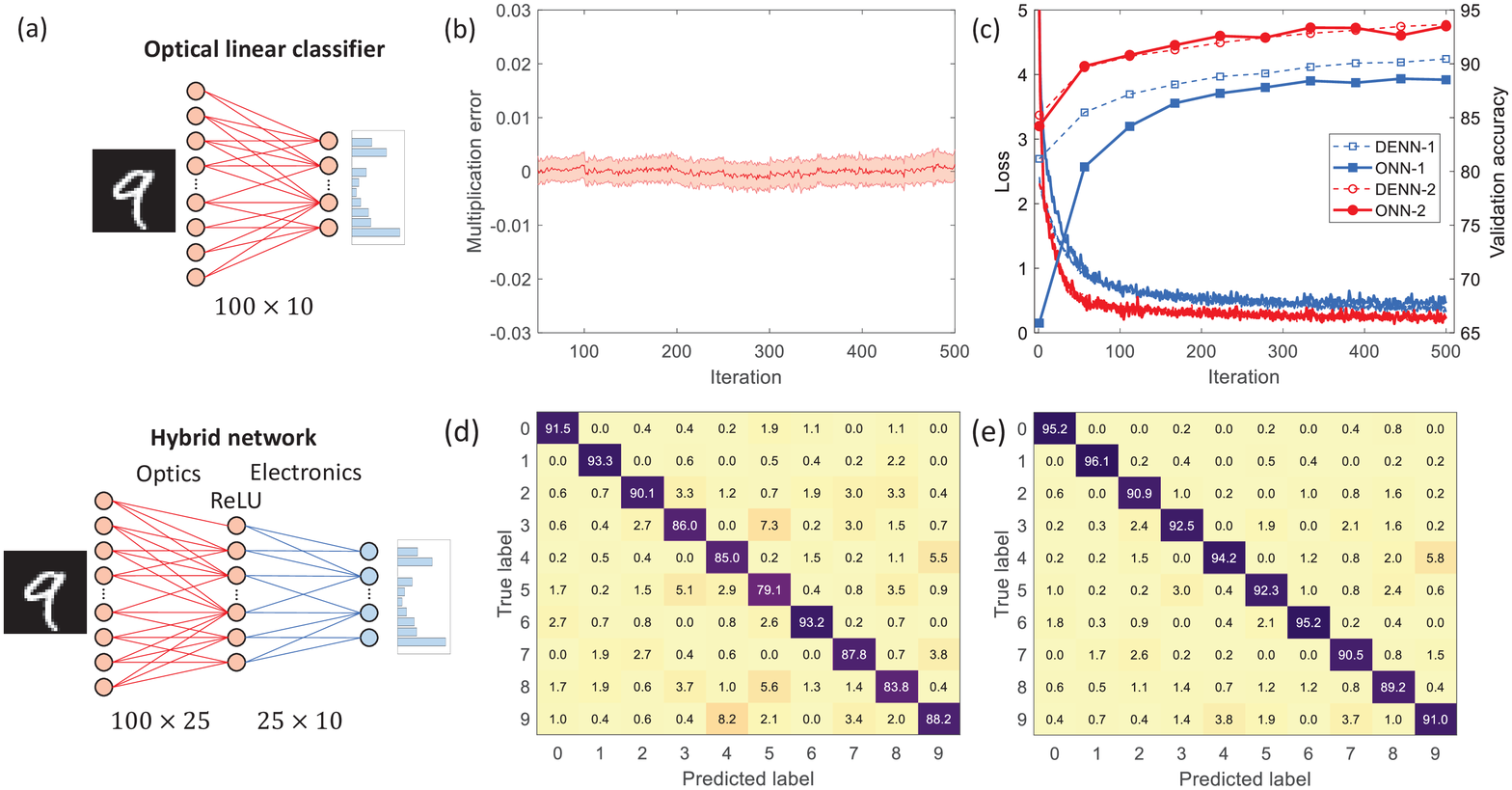}
		\caption{\textbf{Hybrid training of an optical linear classifier (ONN-1) and a hybrid opto-electronic network (ONN-2).} (a, top) Network architecture of ONN-1 with 100 input neurons and 10 output neurons. (a, bottom) Network architecture of ONN-2 with an optical layer, digital ReLU activation and a digital layer. The neuron numbers are 100, 25 and 10 for the input, hidden and output layer. (b) Evolution of the optical MVM error during the training of ONN-1. (c) Learning curves of the hybrid-trained ONNs compared with DENN benchmarks. DENNs have the same network architectures as those of the two ONNs. (d) Confusion matrix of the test set for ONN-1. (e) Confusion matrix of the test set for ONN-2. }
		\label{fig: training}
	\end{figure*}

\begin{figure*}[t]
		\centering
		\includegraphics[width=\textwidth]{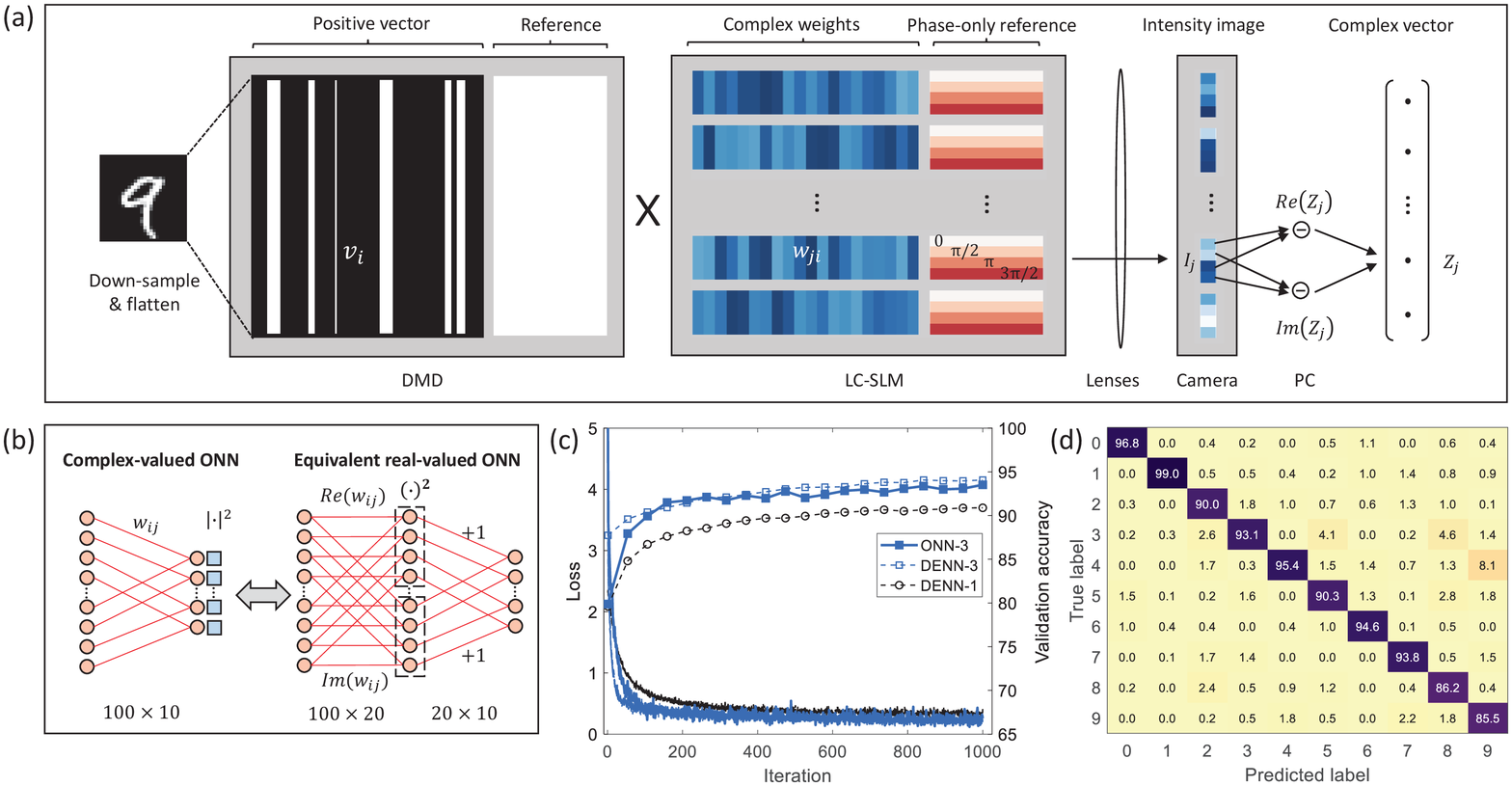}
		\caption{\textbf{Hybrid training of a complex-valued ONN (ONN-3).} (a) Optical encoding and complex-valued measurement scheme. We encode positive vectors on the DMD, complex-valued weights along with phase-only reference on the LC-SLM. At the output we measure intensities and reconstruct the complex values digitally. (b) Network architecture of ONN-3 and the equivalent real-valued ONN. (c) Learning curve of the hybrid-trained ONN-3 compared with DENN-3 with the same network architecture. (d) Confusion matrix of the test set for ONN-3.}
		\label{fig: CV-ONN}
\end{figure*}

We first characterize our optical multiplier by running a large number of real-valued MVMs with matrix sizes of $100\times10$ and $100\times25$, and the results are shown in Fig.~\ref{fig: multiplier}(a) and (b). In these measurements, the input vector and weight matrix elements are random, approximately normally distributed around zero with a standard deviation of 0.5, which is consistent with typical distributions in neural networks. The output vector is normalized by the maximum possible output which is obtained when all the vector and matrix elements are at maximum. The experimentally measured data fall along the diagonal theoretical line. The root-mean-square error (RMSE) for the two matrix sizes are $0.0024$ and $0.0036$, normalized to the output vector  range $[-1, 1]$. %\textcolor{black}{These RMSE levels correspond to about 5-bit precision with $68\%$ confidence ($1\sigma$) over the narrow output range $[-0.1, 0.1]$. We characterize the multiplier in this narrow range because most MVM outputs distribute around zero during the training of ONNs. If we perform MVMs with outputs over the full range $[-1, 1]$, 8-bit precision over the full range may be achieved, because the RMSE of the system is independent of output values, as we reported previously~\cite{spall2020fully}.}

%These RMSE levels then correspond to 8-bit precision with $68\%$ confidence ($1\sigma$) over the full range $[-1,1]$, and are at least 5 times better than those previously achieved~\cite{spall2020fully}. %This should be compared with the mean square value of the MVM vector element% (see Methods for details).

Our optical multiplier also naturally supports complex-valued operation, therefore we next perform complex-valued MVM with the matrix size of $100\times10$, and measure both the real and imaginary parts of the output by setting different LO phases on the LC-SLM. \textcolor{black}{In this work, the inputs encoded by the DMD are always positive-valued, while the weights and outputs are real- or complex-valued. However, complex-valued inputs could also be generated by modulating a binary grating on the DMD~\cite{lee1979binary}.} Fig.~\ref{fig: multiplier}(c) plots examples of the experimentally measured result on the complex plane, compared with theoretical target values. Fig.~\ref{fig: multiplier}(d) shows the complex-valued MVM precision for the real and imaginary parts, and the RMSE values are $0.0034$ and $0.0039$, respectively. 

The precise optical multiplier forms the basis of our ONN, and it can directly work as an optical linear classifier (ONN-1), as shown in Fig.~\ref{fig: training}(a, top). We perform hybrid training of ONN-1 on the MNIST handwritten digits dataset. The dataset consists of 70000 greyscale bitmaps of size $28\times28$. These images are downsampled to size $10\times10$, and then flattened into vectors to be fed to the input layer of the optical linear classifier. At the output layer, 10 neuron values are measured by the camera and used to calculate a digital cross-entropy loss function after applying a digital softmax activation. The target values of the output neurons are 0 or 1 dependent on the label of the bitmap (``one-hot encoding''). We split the entire dataset into training, validation and test sets, each consisting of 60000, 5000, and 5000 bitmaps. The training is implemented in 500  randomly sampled mini-batches with a mini-batch size of 240. \textcolor{black}{Backpropagation was performed digitally with Adam optimizer, with $\beta_1=0.9$ and $\beta_2=0.999$, and an initial learning rate of 0.01. The weight matrices are initialised randomly from a normal distribution $N(0,0.5)$.} 

It is essential that the system errors do not accumulate and blow up during the hybrid training. In Fig.~\ref{fig: training}(b) we plot the evolution of the optical MVM error during the entire training process, and it is clear that our optical system maintains the precision steadily. Fig.~\ref{fig: training}(c) shows the learning curves during the hybrid training, and we see that both the loss function and validation accuracy converge quickly after the first few iterations. The learning curve of a similar digital electronic linear classifier (DENN-1) is also presented for comparison. \textcolor{black}{The DENN is trained with the same hyperparameters and number of iterations.} After the hybrid training, we perform image classification on the test set, and the confusion matrix is shown in Fig.~\ref{fig: training}(d). \textcolor{black}{The classification accuracy reaches $88.0\%$. In comparison, DENN-1 scores $89.9\%$ after training with the same number of iterations.} The slightly slower convergence and lower accuracy of ONN-1 as compared to DENN-1 is mainly due to random dynamic experimental noise, as will be discussed later. We emphasize that during the hybrid training, we apply the standard error backpropagation algorithm without any modeling of the optical system or inclusion of noise.

We next build a more complicated hybrid opto-electronic network (ONN-2) consisting of one optical input layer, one hidden layer with digital ReLU activation and one digital output layer with 100, 25 and 10 neurons at each layer, as shown in Fig.~\ref{fig: training}(a, bottom). Such hybrid networks may simplify the acquisition and processing of data in deep optics and IoT applications by extracting salient features from the optical front-end~\cite{wetzstein2020inference}. Fig.~\ref{fig: training}(c) plots the learning curve of this ONN in comparison with a similar digital electronic network (DENN-2). Fig.~\ref{fig: training}(e) shows the confusion matrix of the test set for ONN-2. \textcolor{black}{The test accuracy reaches $92.7\%$, above the benchmark level of a linear classifier and close to the performance of DENN-2 at $93.2\%$.} 

%As an illustration, we simulate a larger network with 784, 256 and 10 neurons at each layer, and during the training we include random dynamic noise that matches our experimental noise level. In this case the network can reach $94.5\%$ accuracy.

\begin{table*}[ht]
\centering
{
{
\begin{tabular}{ |p{0.45\linewidth} | p {0.15\linewidth} | p{0.15\linewidth} | p{0.1\linewidth} | p{0.1\linewidth}| }
\hline
ONN type & Nonlinearity & Network size & Hybrid training accuracy & DENN accuracy  \\
\hline
 Optical linear classifier (ONN-1) & None & $100\times10$  & 88.0$\%$ & \textcolor{black}{89.9$\%$ (91.8$\%$)} \\
 Hybrid opto-electronic network (ONN-2) & ReLU & $100\times25\times10$ & 92.7$\%$ & \textcolor{black}{93.2$\%$ (95.7$\%$)}   \\
 Complex-valued ONN (ONN-3) & Modulus square & $100\times10$  & 92.6$\%$ & \textcolor{black}{93.8$\%$ (94.8$\%$)}  \\
\hline
\end{tabular}
}
\caption{\textbf{List of ONNs demonstrated in this work.}}\label{Table: List of ONNs}
}\end{table*}

\begin{figure*}[hbt]
    \centering
    \includegraphics[width=\textwidth]{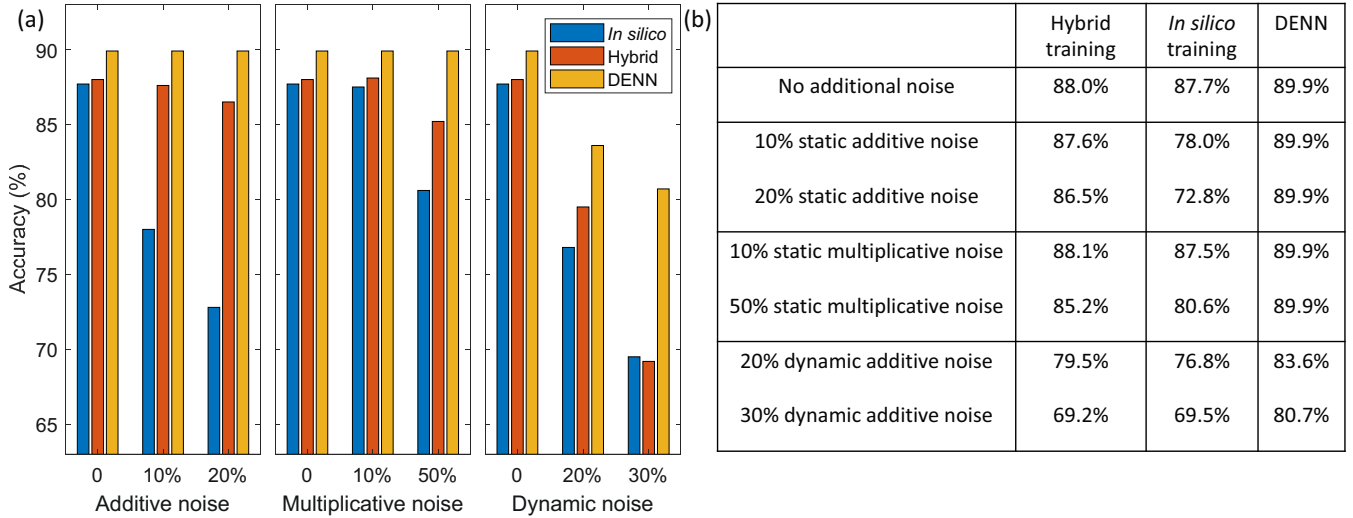}
    \caption{\textbf{Impact of noise during network training.} (a) Comparison of hybrid training and \emph{in silico} training of ONNs and DENN in presence of experimental imperfection. The left, middle and right panel shows effect of static additive noise, static multiplicative noise, and dynamic additive noise. \textcolor{black}{DENN accuracy shown in brackets are obtained after training for twice the number of iterations.} (b) List of accuracy at different noise levels. \textcolor{black}{The noise level is defined as the value of $\sigma$ as defined in the text.}}
    \label{Table: ONN with noise}
\end{figure*}

\subsection{Complex-valued ONN}

It has been recently observed \cite{zhou2021large, li2021spectrally} that diffractive neural networks employing complex-valued operations can outperform linear classifiers, even though the diffractive connections are entirely linear. This is because the intensity detection at the output layer of the complex-valued ONN is equivalent to creating a hidden layer with a square nonlinearity. Consider a single-layer complex-valued optical linear classifier with real-valued inputs $E_{i}$ and complex-valued weights ${w}_{ji}$. At the output layer we detect the intensity of each output unit:
\begin{equation}\label{eq: CV-MVM}
    I_{j} =\left|\sum_{i}{w}_{ji}E_{i}\right|^2 =\left(\sum_{i}\Re({w}_{ji})E_{i}\right)^2 + \left(\sum_{i}\Im({w}_{ji})E_{i}\right)^2.
\end{equation}
We see that this is equivalent to a two-layer real-valued ONN with square activation at the hidden layer, followed by a weight matrix with the fixed values of 0 and 1, connecting  each output neuron to exactly two hidden neurons. This equivalence is depicted in Fig.~\ref{fig: CV-ONN}(b). 

As previously stated, our optical multiplier naturally supports both real-valued and complex-valued operations. We can therefore build a complex-valued ONN (ONN-3) with stronger learning capabilities. A complex-valued ONN was recently built on a photonic integrated circuit, but only with a few neurons per layer~\cite{zhang2021optical}. In comparison, our network size is an order of magnitude larger.

Even though the ONN-3 output is a set of intensities, the complex-valued output neuron amplitudes are required for the calculation of the weight matrix update. %Therefore, we cannot directly perform intensity measurement of the output neurons. 
To measure these amplitudes, we change the relative phase $\phi$ between the LO field and the signal, as shown in Fig.~\ref{fig: CV-ONN}(a). The real parts are extracted by setting $\phi=0$ and $\phi=\pi$ and subtracting the measured intensities, and the imaginary parts are extracted similarly by setting $\phi=\pi/2$ and $\phi=3\pi/2$. Instead of setting these four phases sequentially, we apply four spatially-separated reference blocks of different phases for each output unit. That is to say, each output unit is spatially split into four regions yielding the four required interference results. Therefore, the real and imaginary parts of the output can be read out from one single camera frame. Upon readout, the modulus square activation is computed digitally. To complete the hybrid training, digital error backpropagation can be run with the ONN modelled as the equivalent two-layer real-valued network.

The learning curve in Fig.~\ref{fig: CV-ONN}(c) converges quickly, with the highest classification accuracy of $93.6\%$ achieved with the validation set, and  $92.6\%$ with the test set. We can also carry out \emph{in silico} training and perform intensity measurement directly, by simply switching off the LO field. In this case we obtain a classification accuracy of $92.2\%$ with the test set.

% \begin{table*}[t]
% \centering
% {{
% \begin{tabular}{ |c|c|c|c| }
% \textcolor{black}ine
% & Hybrid training & \emph{In silico} training & DENN  \\
% \textcolor{black}ine
%  No additional noise & 88.0$\%$  & 87.7$\%$ & 91.8$\%$ \\
%  10$\%$ static additive noise & 87.6$\%$ & 78.0$\%$  & 91.8$\%$ \\
%  20$\%$ static additive noise & 86.5$\%$  & 72.8$\%$  & 91.8$\%$ \\
%  20$\%$ static multiplicative noise & 88.1$\%$ & 87.5$\%$ & 91.8$\%$ \\
%  50$\%$ static multiplicative noise & 85.2$\%$ & 80.6$\%$ & 91.8$\%$ \\
%  20$\%$  dynamic additive noise & 79.5$\%$ & 76.8$\%$ & 81.7$\%$ \\
%  30$\%$  dynamic additive noise & 69.2$\%$ & 69.5$\%$ & 79.7$\%$ \\
% \textcolor{black}ine
% \end{tabular}
% }
% \caption{\textbf{Comparison of hybrid training and \emph{in silico} training in presence of experimental imperfection.}}\label{Table: ONN with noise}
% }\end{table*}

\subsection{Impact of noise}

We have demonstrated the efficacy of our hybrid training scheme with three different types of ONNs, as summarized in Table~\ref{Table: List of ONNs}. We now use ONN-1 to explore the performance of hybrid training compared to traditional \emph{in silico} training in different noisy environments. 

In order to systematically compare different types of noise, we start with a well-controlled, low-noise environment. After carefully calibrating our system and performing the \emph{in silico} training, ONN-1 reaches $87.7\%$ classification accuracy, nearly the same as that of the hybrid training. This indicates that we have eliminated most aberrations and systematic errors, which is consistent with our small RMSE of the optical multiplier.

% We now explore the performance of hybrid training in different noisy environments, as compared to traditional in-silico training, using ONN-1. 

% Different physical systems exhibit experimental imperfection and noise of different origins, therefore it is important to perform the comparison in a well-controlled environment. 

%We emphasize that during this in-silico training we don't model the physical system, and we don't introduce any artificial noises to force the network to be robust against experimental imperfections.

In the comparative study, we introduce different imperfections to the optical setup via the LC-SLM, and the results are shown in Fig.~\ref{Table: ONN with noise}. The first imperfection is static additive noise: a random bias $w_{ji}\rightarrow w_{ji}+\epsilon_{ji}, \epsilon\in N(0, \sigma)$ applied to each weight matrix element, which remains unchanged during the training and testing. This can arise from ambient light, imprecise device calibration, etc. In our experiment,  we randomly sample the bias which is fixed during the entire training and testing process. %The bias $\epsilon_{ji}$ is added to each weight element individually during the weight update: . 
As seen from Fig.~\ref{Table: ONN with noise}(a), hybrid training is robust to such static additive noise, while the accuracy of \emph{in silico} training drops to $72.8\%$ at $20\%$ noise level \textcolor{black}{($\sigma=0.2$)}. 

A second common imperfection is static multiplicative noise $w_{ji}\rightarrow w_{ji}\times \eta_{ji}, \eta\in N(1, \sigma)$. This may be caused by non-uniform transmission of different optical channels, imperfect interference, different responses of photodetectors, etc.  From Fig.~\ref{Table: ONN with noise}, we see that hybrid training is also robust against such  noise, while the performance of \emph{in-silico} training degrades to $80.6\%$ at $50\%$ noise level.

The last major type of imperfection is dynamic noise, which fluctuates over time. This may arise from imprecise device calibration, environmental fluctuation, etc. In the experiment we model the dynamic noise by additive noise (as defined above) applied to each weight element, which is randomly re-sampled at each weight update. 
%The noise level is defined similarly as the ratio between noise standard deviation and signal standard deviation. 
Our results show that the ONN trained in either the hybrid or \emph{in silico} scheme is sensitive to such random dynamic noise, and the accuracy drops to about $69\%$ at $30\%$ noise level, i.e.~at the similar level as an equivalent DENN. \textcolor{black}{In contrast to static errors, the cancellation of dynamic noise by hybrid training is not possible because the errors change from one iteration to the next.}% \textcolor{black}{This is because the fluctuating noise limits the precision of each gradient descent step and the error can not be systematically compensated by the weight update.}
%\textcolor{black}{We emphasize that this doesn't mean ONNs are particularly fragile. In fact, even the DENN degrades significantly with the addition of random dynamic noise, as shown in Table~\ref{Table: ONN with noise}. } 
%From our study, it can be concluded that hybrid training can help close the reality gap in systems exhibiting significant static noise. 

\textcolor{black}{We believe that the difference in the ONN and DENN performance that we see in Table~\ref{Table: List of ONNs} arises due to dynamic noise-like effects in the ONN. Specifically, the SLM pixel cross-talk and diffraction result in deviation of the effective weight matrix elements from the desired values, with these deviations being different in every iteration. To mitigate these effects, every 50 iterations, we re-calibrate the system by running a few MVM examples with known outputs and adjust the parameters that map the camera grey level to MVM output. In the future, in order to increase the accuracy of our ONNs and extend our method to more complicated networks, more elaborate error correction methods would be required. }

\begin{figure}[t]
		\centering
		\includegraphics[width=0.45\textwidth]{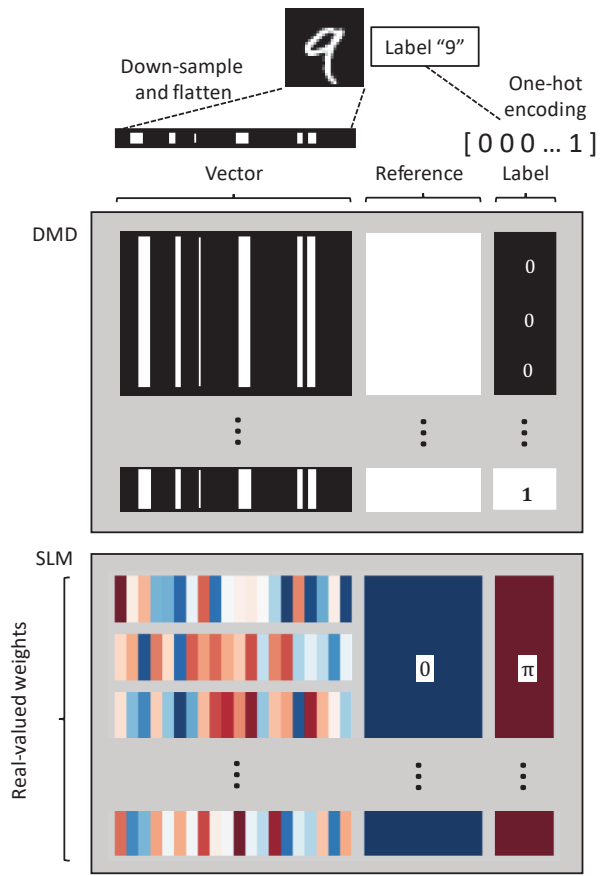}
		\caption{\textbf{Encoding scheme for optical calculation of the error vector in ONN-1.} The destructive interference between the label region and signal region yields the optical calculation of the error vector when we use MSE loss function.}
		\label{fig: ONN-1s}
\end{figure}

\subsection{Optical calculation of the error vector}

There is a long-standing challenge to implement all-optical ONN training ~\cite{brunner2021competitive}, such that the error terms are calculated and backpropagated optically. Our system can be modified to demonstrate an important step towards realising this goal. To this end, we replace the categorical cross-entropy loss function in ONN-1 with mean-squared error (MSE) loss function,
\begin{equation}
\Loss = \frac{1}{2} \left( z_j^{(1)} - y_j \right)^2,
\label{eq:mse_loss}
\end{equation}
where $z_j^{(1)}$ is the output of the linear network and $y_j$ is the corresponding one-hot encoded label. With this  network architecture, the error vector to be calculated is
\begin{equation}
\delta_j = z_j^{(1)} - y_j.
\label{eq:mse_gradient}
\end{equation}
This is easy to implement optically, by destructive interference between the ONN output and an optically-encoded label, and the schematic is shown in Fig.~\ref{fig: ONN-1s}. We introduce a third active region on both the DMD and LC-SLM, which we call the label region. During hybrid training, in addition to encoding the input vector, we encode the target label on the DMD, and the label region on the LC-SLM is set to maximum reflection, with a $\pi$ phase shift relative to the reference region. After passing through the cylindrical lens, the label destructively interferes with both the signal and LO fields at the output plane. Intensity measurement therefore directly yields the error term (\ref{eq:mse_gradient}). This optically-calculated error is then processed digitally to update the weights.

In this hybrid training setting we achieve a peak validation accuracy of $83.3\%$. We then perform inference on the test set, by switching off the label region and measuring only the network output, and achieve an accuracy of $83.4\%$. By comparison, simulating this ONN with a DENN of the same architecture and MSE loss function, yields a test accuracy of $83.6\%$. \textcolor{black}{The lower accuracy as compared to that of ONN-1 or DENN-1 is because the MSE loss function is not well suited to classification tasks. We can expect better performance with optical calculation of the error vector when solving regression problems where MSE is more suitable.} 

% Note that we don't have any nonlinearity in this system, and the accuracy level can be significantly improved by introducing a nonlinear activation layer.

\section{Discussion}

Our ONNs support up to 100 neurons per layer, limited by the resolution and the chosen grating period of the LC-SLM, as well as our matrix encoding method. %Therefore, the maximum matrix size is determined by the resolution of LC-SLM. 
Our LC-SLM model has a resolution of $1440\times1050$, of which we use $1140\times1050$ pixels as the signal and $300\times1050$ pixels as the reference region. We use a diagonal blazed grating with horizontal and vertical periods of 10 pixels, so our maximum matrix size is about $110\times100$. \textcolor{black}{We found the precision of our MVM to be largely independent of the input vector dimension within these limits.} Larger network size can be achieved by using LC-SLM models with higher resolution and reducing the grating period. Up to 1000 neurons per layer should be within reach. 

%\textcolor{black}{In our system, we prepare 4-bit inputs with the DMD, encode weight matrices with a 10-bit LC-SLM, and measure the output with an 8-bit camera. With these devices, 5-bit real-valued and complex-valued optical MVMs are achieved and used to support three different ONNs.}

\textcolor{black}{With our hybrid training scheme, we reach similar classification accuracy as that of DENNs, and even higher accuracy can be achieved with longer training times. As an illustration, the three DENNs trained with twice the number of iterations converge at $91.8\%$, $95.7\%$ and $94.8\%$, as shown in Table~\ref{Table: List of ONNs}. Therefore, future ONNs with long-term system stability can be expected to improve performance. A second important limiting factor of the classification accuracy is the dynamic noise and imperfect device calibration that leads to weight-dependent fluctuations, such as SLM pixel cross-talk. These imperfections cannot be corrected during the training, and need to be suppressed to improve network performance. A final limitation is that the weights are constrained in passive ONNs. In our system we chose to clip any weights above the bound, and although this leads to a reduced search space during the training, it did not significantly affect the network performance in this work.}

During the hybrid training, our computation speed is limited by the frame rates of the DMD, LC-SLM and camera. In our chosen mode of operation, \textcolor{black}{the DMD works at 1440 Hz frame rate in the binary mode}, while our camera works at a maximum frame rate of 1480 Hz. The LC-SLM only needs to update once per mini-batch, i.e. at 6 Hz for a mini-batch size of 240 images. Therefore its maximum operational refresh rate of 60 Hz supports up to 14.4 kHz DMD frame rate with this mini-batch size. Therefore, our system frame rate is limited by the DMD at 1440 Hz, and the computation speed is $1440\times100\times25\times2=7.2\times10^{6}$ operations per second (where 100 and 25 are the first ONN layer dimensions). Today's advanced DMD and LC-SLM models support a maximum frame rate of up to 20 kHz and 1 kHz respectively, and one can replace the camera by an ultra-fast photodetector array. Therefore, the system frame rate can be increased by at least 10 times. Assuming a two-layer ONN with 1000 neurons per layer that updates at 20 kHz, the computation speed would be $4\times10^{10}$ operations per second. Although ONNs with similar or even higher computing rates have been demonstrated~\cite{xu202111, feldmann2021parallel, miscuglio2020massively}, these demonstrations are limited to convolutional architectures. \textcolor{black}{In contrast, our ONNs are fully connected, can be readily scaled further, and are the only ones capable of rapid update.}

%In contrast, our ONNs are fully connected, and in this domain, to our knowledge, they have the largest layer sizes demonstrated to date~\cite{shen2017deep, zuo2019all, zhang2021optical} and are the only ones capable of rapid update. 

Instead of using a DMD as the input modulator, one can also use an electro-optical modulator array with a bandwidth that can exceed $10$ GHz ~\cite{miller2017attojoule, hamerly2019large}. Such a two-layer ONN with 100 input neurons and 1000 neurons in subsequent layers would reach a computation speed of $2\times10^{15}$ operations per second with a power efficiency that is a fraction of the same computational power provided by a cluster of GPUs.

Our work shows that analog systems with limited signal-to-noise ratio can still be physically trained to reach high performance, and this is a crucial step towards the more advanced goal of all-optical training of neural networks. We demonstrate a further step towards this goal by modifying our ONN  to allow optical calculation of the error vector. The final remaining challenge is optical backpropagation in a deeper network with optical nonlinearity.     

\medskip

\noindent\textbf{Acknowledgements}
This work is supported by an EPSRC IAA award. X.G. acknowledges support from the Royal Commission for the Exhibition of 1851 Research Fellowship.

\medskip

\noindent\textbf{Author contributions} 
J.S. and X.G. carried out the experiment and performed the data analysis. All the authors jointly prepared the manuscript. This work was done under the supervision of A.L. 

\medskip

\noindent\textbf{Disclosures} J.S., X.G. and A.L. are inventors of patent application GB 2203480.5 on method and apparatus for hybrid training of optical neural networks.

\medskip

\noindent\textbf{Conflict of interest} The authors declare no conflicts of interest.

\medskip

\bibliography{HybridTraining}

\end{document}